%% file: main.tex
\documentclass[times,twocolumn,final,authoryear]{elsarticle}

\usepackage{ycviu}
\usepackage{framed,multirow}

\usepackage{tikz}
\usepackage{comment}
\usepackage{amsmath,amssymb} 
\usepackage{booktabs}
\usepackage{color}
\usepackage{array}
\usepackage{subcaption}
\usepackage{soul}

\usepackage[accsupp]{axessibility}  
\usepackage{pifont}
\newcommand{\cmark}{\ding{51}}%
\newcommand{\xmark}{\ding{55}}%

\usepackage{amssymb}
\usepackage{latexsym}

\usepackage{url}
\usepackage[colorlinks=true]{hyperref}
\usepackage{xcolor}
\definecolor{newcolor}{rgb}{.8,.349,.1}

\mathchardef\mhyphen="2D
\newcommand{\methodname}{I-Paint}
\newcommand{\elia}[1]{#1}
\newcommand{\willi}[1]{#1}
\newcommand{\new}[1]{\textcolor{black}{{#1}}}
\renewcommand{\st}[1]{\unskip}

\input{paper_sections/notation}

\journal{Computer Vision and Image Understanding}

\begin{document}
\begin{frontmatter}

\title{Interactive Neural Painting}

\author[1]{Elia \snm{Peruzzo}\corref{cor1}}
\author[1]{Willi \snm{Menapace}}
\author[2]{Vidit \snm{Goel}}
\author[3]{Federica \snm{Arrigoni}}
\author[4]{Hao \snm{Tang}}
\author[5]{Xingqian \snm{Xu}}
\author[2]{Arman \snm{Chopikyan}}
\author[2]{Nikita \snm{Orlov}}
\author[2]{Yuxiao \snm{Hu}}
\author[2,5]{Humphrey \snm{Shi}}
\author[1]{Nicu \snm{Sebe}}
\author[1,6]{Elisa \snm{Ricci}}

\address[1]{University of Trento, Italy}
\address[2]{Picsart AI Research (PAIR)}
\address[3]{Politecnico di Milano, Italy}
\address[4]{ETZ, Switzerland}
\address[5]{University of Illinois Urbana Champaign (UIUC), US}
\address[6]{Fondazione Bruno Kessler, Italy}

\received{1 May 2013}
\finalform{10 May 2013}
\accepted{13 May 2013}
\availableonline{15 May 2013}
\communicated{S. Sarkar}

\begin{abstract}
In the last few years, Neural Painting (NP) techniques became capable of producing extremely realistic artworks. 
This paper advances the state of the art in this emerging research domain by proposing the first approach for Interactive NP. Considering a setting where a user looks at a scene and tries to reproduce it on a painting,
our objective is to develop a computational framework to assist the user’s creativity by suggesting the next strokes to paint, that can be possibly used to complete the artwork. To accomplish such a task, we propose \methodname, a novel method based on a conditional transformer Variational AutoEncoder (VAE) architecture with a two-stage decoder.
To evaluate the proposed approach and stimulate research in this area, we also introduce two novel datasets. 
Our experiments show that our approach provides good stroke suggestions and compares favorably to the state of the art. Additional details, code and examples are available at \href{https://helia95.github.io/inp-website}{the project website}.
\end{abstract}

\begin{keyword}
\MSC 41A05\sep 41A10\sep 65D05\sep 65D17
\KWD Keyword1\sep Keyword2\sep Keyword3

\end{keyword}

\end{frontmatter}


\input{paper_sections/main_introduction}
\input{paper_sections/main_related}

\input{paper_sections/main_method}
\input{paper_sections/main_experiments}
\input{paper_sections/main_conclusions}

\section*{Acknowledgement}
\noindent The research was supported by the MUR PNRR project FAIR - Future AI Research (PE00000013) funded by the NextGenerationEU and the the PRIN project CREATIVE (Prot. 2020ZSL9F9).

\bibliographystyle{model2-names}
\bibliography{refs}

\newpage
\appendix
\section*{Appendix}
\input{paper_sections/appendix}
\end{document}

%% file: paper_sections/notation.tex
\newcommand{\fvisual}{f_{c,\text{visual}}}
\newcommand{\fstrokes}{f_{\text{strokes}}}
\newcommand{\fdim}{\mathrm{d_{emb}}}

%% file: paper_sections/main_introduction.tex
\begin{figure*}[t]
\centering
\includegraphics{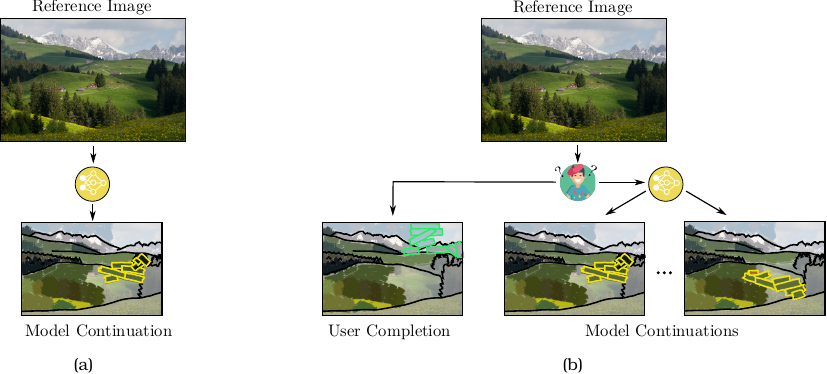} 
\caption{\elia{Inspired by en-plain-air setting, we tackle the task of representing a reference image on the canvas. The final painted result is obtained by sequentially placing strokes on the canvas, refining the artwork until the user is satisfied. In this figure we describe a single step of such a process. In Neural Painting (a),
a deep architecture is learned in order to create a realistic artwork of a reference image with a sequence of strokes (outlined in yellow). Note that \textbf{the user is not involved} in this process. 
Differently, in Interactive Neural Painting (b), \textbf{the user directly contributes} to the generation process in an interactive loop with the deep model. At each iteration, the model provides a set of stroke suggestions based on the reference image and current state of the painting, and the user either selects a suggested set of strokes (outlined in yellow) or directly draws new strokes on the canvas (outlined in green).}
}
\label{fig:teaser}
\end{figure*}

\section{Introduction}

One of the main objectives of image generation methods is enabling novel and more powerful ways in which humans can express their creativity. This objective inspired a lot of research and advancements in deep generative models, that are now able to produce outputs with photorealistic quality in several generation tasks. A recent trend in deep image generation is that of improving the way in which users can control  and interact with the generation process, thus providing tools to convey users intentions.
In this context, recent works allow users to generate or edit images with high quality by sketching (\cite{ghosh2019interactive,liu2021deflocnet}), modifying the semantic layouts (\cite{lee2020maskgan,ling2021editgan,park2019gaugan,zhu2020sean}), or providing a text prompt (\cite{bau2021paint,nichol2021glide,ramesh2021zero,xu2021predict}).
These methods, however, allow users to influence the final output only in an indirect manner, \textit{i.e.} through the sketched semantic layout or the input text.

Recently, several learning-based methods for painting generation have been proposed, commonly referred to as Neural Painting (NP) methods.
Differently from other generative approaches that operates in the pixel space, NP methods leverage a parameterized brushstroke representation which is more aligned to how humans  visualize and conceptualize an artwork (\cite{Kotovenko_2021_CVPR,liu2021painttransformer,zou2021stylized}).
The strokes-based vector representation offers several benefits compared to the pixel-based representation, such as the ability to modify or erase individual strokes. Additionally, separating the representation from the rendering process enables the strokes to be rendered at any desired output resolution.

Painting has historically been a powerful tool with which humans expressed their creativity. However, in this respect, current NP methods are inherently limited, as they are only designed to reconstruct and stylize a given target image, leaving no possibility for the user to influence the generation process.
Lacking the ability to integrate users' painting style, these methods are unsuitable in interactive scenario.
This work represents the first attempt to fill this gap in the literature \willi{and bring the next level of interaction to NP}.
Inspired by \textit{en plein air} painting, \textit{i.e.} the setting where a painter looks at an outdoor scene and tries to represent it on a canvas, we introduce the novel task of Interactive Neural Painting (see Fig.~\ref{fig:teaser}). Specifically, we propose an iterative and interactive process where, given a reference image the user would like to paint and an incomplete canvas, {a computational tool} {based on NP techniques} assists the user in drawing the painting. The tool provides multiple suggestions about the next strokes at each iteration, from which the user can choose to continue its artwork.
Such a tool speeds up the painting process but, differently from existing NP approaches, leaves the user a high degree of control on the final output, with the potential of making painting an artistic medium accessible not only to highly-skilled individuals. Our system can be integrated into digital drawing tools used by amateur and professional artists, such as Adobe Photoshop, GIMP, and Krita, as shown by our demo in the \emph{Supp. Mat.}.

We devise the first method for Interactive NP (INP). Our method, which we call \methodname, introduces a conditional transformer VAE architecture that generates stroke suggestions.
To ensure seamless interaction with the user, our method is specifically trained to produce stroke suggestions that closely match the dynamics of the painting process represented in a given dataset of painting demonstrations. 
\new{The dataset is built to reflect in a synthetic manner the main aspects valued by a human painter such as color consistency, local proximity and object-based painting. Additionally, artists typically begin by portraying a rough depiction of the reference image, and incrementally incorporate finer details during the painting process (\citet{zhao2020painting, Singh2021IntelliPaintTD}). We follow previous work (\cite{zou2021stylized, liu2021painttransformer}) and adopt a coarse-to-fine assumption in our dataset to reflect this behaviour, with rough strokes spatially covering the  canvas in the first stages of the painting an detailed localized strokes towards the final stages (see Sec.~\ref{sec:dataset} for more details).} 
To effectively learn the characteristics of the stroke dataset, we introduce a distribution matching loss that minimizes the discrepancies between the suggested strokes and the painting demonstrations. In addition, a two-stage VAE decoder is proposed that tightly integrates visual features into the stroke prediction process. 
Furthermore, we make our approach probabilistic by nature to capture the complex distribution of possible continuations given the current canvas state. In this way, \methodname~can produce multiple suggestions about what to paint next.
We demonstrate our method on two novel datasets which we specifically introduce for the INP task, built upon the \textit{ADE 20K Outdoor}~\cite{zhou2017ade20k} and \textit{Oxford-IIIT Pet}~\cite{parkhi2012cats} datasets. Our extensive evaluation shows that our model produces a wide set of suggestions that closely match the characteristics of the painting demonstrations. Quantitative comparison against state of the art NP methods, supported by results on a user study, demonstrates state-of-the-art performance of our method.

\noindent \textbf{Contributions.} To summarize, our main contributions are: 

\begin{itemize}
    \item The novel image generation task of INP,  which for the first time brings interactivity to neural painting.
    \item The first approach based on conditional transformer VAE to address this task, with specific architectural choices and training protocols.
    \item Two novel synthetic datasets and a set of evaluation metrics for training and evaluating INP models, to foster and assess the research in this new area.  
\end{itemize}

%% file: paper_sections/main_related.tex
\section{Related Work}
\label{sec_related}
In this section, we discuss the most related works in the field of NP and interactive image generation.

\subsection{Neural Painting}
Neural Painting techniques are derived from the intriguing idea of teaching machines how to paint. 
NP is typically formalized as the process of artistically recreating a given image using a neural network which generates a series of strokes. Several approaches are present in the literature that address this task.
Some of them make use of reinforcement learning (RL) \cite{Huang2019LearningTP,SchaldenbrandO21,Singh2021IntelliPaintTD,Singh_2021_CVPR,Xie2012ArtistAA}, where, given the current environment represented by the present status of the canvas and a reference image, an agent is trained to predict the parameters of the next strokes. The training objective is formulated as the maximization of the cumulative rewards of the whole painting process, typically expressed as the increase in similarity between the new canvas state and the reference image. Since no gradients need to be directly backpropagated from the reward function, RL-based methods do not require a differentiable stroke rendering procedure. \\ 

Other methods, instead, make use of a differentiable stroke renderer that allows direct optimization of a loss objective. Among these methods, \cite{zou2021stylized} and \cite{Kotovenko_2021_CVPR} directly optimize a set of parameters describing the stroke sequence, producing high-quality results at the cost of long inference times. Other works overcome this limitation by using a model to predict stroke parameters rather than directly optimizing them. In this context, the state of the art is represented by Paint Transformer (PT,  \cite{liu2021painttransformer}), where NP is expressed as a set prediction problem and a transformer-based architecture is proposed that predicts the parameters of a stroke set with a feedforward network. Our method shares similarities with \cite{liu2021painttransformer} as we also assume a differentiable stroke renderer and predict stroke parameters with a feedforward network. However, we address a different (and new) task, by focusing on an \emph{interactive} setting \willi{that requires seamless integration between model predictions and user inputs}. \\ 

Finally, a different class of methods focuses on the closely related task of sketch generation which consists in the generation of \willi{abstract} sketches. Sketch-RNN (\cite{Ha_sketchRNN}) and Sketch-BERT (\cite{lin2020sketchbert}) represent sketches as sequences of points and are based respectively on RNN and transformer models. While these methods can be employed in interactive tasks such as sketch completion, they are not able to reproduce natural images and do not model realistic painting effects.

\subsection{Interactive Image Generation}
Interactive image generation refers to the task of automatically generating photo-realistic images, conditioned on user inputs. 
Early works fall into two directions: (1) image-to-image translation, which investigates the problem of translating an input image to a target domain, allowing to synthesize photos from label maps or reconstruct objects from edge maps (\cite{isola2017image,tang2019multi,zhu2017unpaired,zhu2017toward}); (2) learning a human-interpretable latent space (\cite{chen2016infogan}), projecting a natural image into it, manipulating the latent code to achieve an edit, and synthesizing a new image accordingly (\cite{abdal2019image2stylegan,abdal2020image2stylegan++,BrockLRW17,lin2021anycost,zhu2016generative}). To provide a more compelling experience, recent works on interactive image generation allow more user-friendly interaction, e.g. by means of sketches (\cite{ghosh2019interactive,liu2021deflocnet}), semantic maps (\cite{lee2020maskgan,ling2021editgan,park2019gaugan,tang2020local,zhu2020sean}), \elia{paint strokes (\cite{cheng2022adaptively, singh2022paint2pix})}, and text (\cite{bau2021paint,nichol2021glide,ramesh2021zero,xu2021predict}).  \cite{ghosh2019interactive} introduced iSketchNFill, an interactive GAN-based sketch-to-image translation method that helps novice users to easily create images of simple objects with a sparse sketch and the desired object category.
Differently, GauGAN (\cite{park2019gaugan}) converts a semantic segmentation mask to a photo-realistic image with a spatially-adaptive normalization layer. To flexibly manipulate an existing image, \cite{Bau:Ganpaint:2019} allows the user to perform a localized edit of an image by selecting a specific region, while \cite{liu2021deflocnet} empowers the user to edit low-level details by sketching the desired modifications. 
\elia{\cite{singh2022paint2pix} condition the image generation on strokes painted by the user, to provide a more intuitive way compared to segmentation maps, while \cite{cheng2022adaptively} rely on sketches and paint strokes to guide the generation process, allowing both flexibility and precise control.} Fueled by the success of text-to-image generation (\cite{ramesh2021zero}), very recent works proposed to control image manipulations with natural language, creating an intuitive way of interaction for the user. Notably, \cite{jiang2021language,shi2021learning} focused on the problem of global image editing, while \cite{xia2021tedigan} proposed a unique framework to both generate and manipulate images using text inputs.  \\

However, all the aforementioned methods are evaluated by the quality of the generated results, the diversity of the suggestions, and how closely they match the users’ input. The proposed task of INP adds an additional level of complexity. Since INP gives the users complete stroke-by-stroke control over the final artwork, it is necessary to represent the \emph{process} that leads to the final result.To ensure smooth interaction with the user, the method should follow a paint-like-demonstration behavior (see Sec.~\ref{sec:method} for a discussion). This requirement, and the level of control over the final output, is peculiar to the task of INP, and differentiates it from the existing literature in interactive image generation.

%% file: paper_sections/main_method.tex
\section{Methodology}
\label{sec:method}
In the following we describe our method in detail: Sec. \ref{sec:problem_formulation} provides a formalization of the task and an overview of the proposed method, Sec.~\ref{sec:context_encoder} describes the architectural components, respectively the context encoder, the VAE encoder and decoder, Sec.~\ref{sec:training} illustrates the employed losses and training procedure, Sec.~\ref{sec:inference} describes the inference process, while Sec.~\ref{sec:implementation_details} describes the implementation details.
\subsection{Problem Formulation and Overview}
\label{sec:problem_formulation}

We start the section by formalizing the task of Interactive Neural Painting (INP). We assume a dataset $\mathcal{D}$ of reference images paired with a sequence of stroke parameters $s=s^{1:T}$ of length $T$, representing a decomposition of the image into a sequence of individual strokes. Each stroke is represented \new{by a tuple of eight parameters} as $s=(x, \rho, \sigma, \omega)$, where $x$ is the position of the stroke center on the canvas, $\rho$ is the color, $\sigma$ represents the stroke size expressed as height and width, and $\omega$ is the stroke orientation. At time $t$, given a reference image $I_{\mathrm{ref}}$ and the corresponding sequence of strokes up to the current time $s^{1:t}$, the INP task consists in predicting a set of stroke sequences of length $k$. The set of predicted stroke sequences is presented to the user who can either select one sequence, partially or in its entirety, as the painting continuation or manually define the next strokes if no proposed sequence captures the user's current painting intentions (see Fig.~\ref{fig:teaser}). Note that predicting a sequence of length $k>1$ (\cite{liu2021painttransformer,zou2021stylized}) gives the user the possibility to better understand whether the proposed continuation corresponds to her painting intentions. The operation is repeated iteratively until completion of the painting. \new{The expected behaviour of predicted strokes should exhibit the following characteristics:
\begin{itemize}
    \item Each sequence makes the canvas more similar to the reference image, hence assisting the user in the final goal of completing the painting.
    \item Each sequence presents the same characteristics of the dataset stroke sequences in terms of positioning, color, size and orientation, thus ensuring seamless interaction between the user and the painting agent.
    \item The predicted set contains diverse sequences that cover the main possible continuations of the painting process, hence providing diverse continuation to the user among which to choose. 
\end{itemize}
We devise a set of quantitative evaluation metrics that captures each of these desired behaviours and present them in Sec.~\ref{sec:metrics}.\\
}

\begin{figure*}[!ht]
\centering
\includegraphics[width=\linewidth]{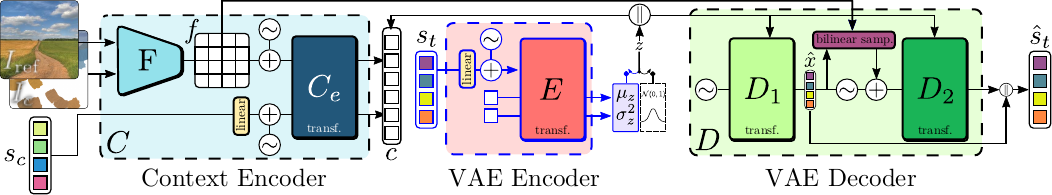}
\caption{\elia{\textbf{Overview of \methodname}. Our model is based on a conditional VAE architecture. A context encoder $C$ extracts a context vector $c$ from the reference and context images ($I_{\mathrm{ref}}, I_{\mathrm{c}}$ respectively) and the context strokes $s_c$. The VAE encoder $E$ encodes the target stroke sequence $s_t$ into a posterior distribution. A latent code $z$ is sampled and used in conjunction with $c$ to condition the VAE decoder that reconstructs the target stroke sequence. \new{At inference time, the latent code $z$ is sampled from the prior distribution $\mathcal{N}(0,1)$. The decoder $D$ is composed of two steps, $D_1$ which predicts the position of the strokes $\hat{x}$, and $D_2$, which predicts the remaining parameters that are concatenated with $\hat{x}$ to form the final prediction $\hat{s}_t$.} Transformers (\cite{vaswani2017attention}) are used to model $C$, $E$ and $D$. We use $\sim$ to represent sinusoidal positional encodings and $\mathbin\Vert$ to represent concatenation. The \textcolor{blue}{blue} outline denotes components that are used only during training.}}
\label{fig:architecture}
\end{figure*}

In this paper we propose \methodname, a method for INP. Our approach consists in a transformer-based conditional VAE architecture and is depicted in Fig.~\ref{fig:architecture}. At time $t$, given a reference image $I_{\mathrm{ref}}$, context strokes $s_c = s^{t-k+1:t}$, and a context image $I_c$ defined as the rendering of strokes $s^{1:t}$, the context encoder $C$ extracts a context vector $c = C(I_{\mathrm{ref}},I_c,s_c)$. During training, the VAE encoder $E$ encodes the target stroke sequence $s_t = s^{t+1:t+k}$ into a posterior gaussian distribution $\mu_z,\sigma^2_z=E(s_t,c)$ and the latent code $z$ is sampled from it. During inference, instead, the latent code is sampled from the prior distribution $\mathcal{N}(0, 1)$.
The latent code $z$ is used in conjunction with $c$ to condition the VAE decoder that produces the sequence of inferred target strokes $\hat{s}_t=D(z,c)$.

\subsection{Architecture Components}
\label{sec:context_encoder}
Next, we describe the architectural components of \methodname~depicted in Fig. \ref{fig:architecture}.

\paragraph{Context Encoder} The context encoder $C$ receives as input the reference image $I_{\mathrm{ref}}$ $\in \mathbb{R}^{3 \times H \times W}$, the context image $I_c$ $\in \mathbb{R}^{3 \times H \times W}$, the sequence of context strokes parameters $s_c$ \new{$\in \mathbb{R}^{k \times 8}$} and produces a representation of the context $c$. 
First, a visual feature encoder $F$, modeled as a CNN, extracts visual features \new{$\fvisual \in \mathbb{R}^{\mathrm{dim} \times H' \times W'}$} 
from the input images. Following \cite{liu2021painttransformer}, we model $F$ as a separate backbone for reference and context images respectively and concatenate the output features along the channel dimension to obtain $\fvisual$.
A linear layer is used to extract context stroke features from $s_c$, resulting in features $\fstrokes \in \mathbb{R}^{k \times \fdim}$. 
Successively, we flatten the spatial dimensions of the visual features $\fvisual \in \mathbb{R}^{(H' \cdot W') \times \fdim}$, and concatenate them with the strokes features $\fstrokes$ along the sequence length dimension. The resulting token sequence $c_{in} \in \mathbb{R}^{L \times \fdim}$, with $L=(H'\cdot W')+k$, is enriched with 3D sinusoidal positional encodings (\cite{vaswani2017attention}) which are added to each element of the sequence. Two encoding dimensions represent the $x$ and $y$ coordinates of the visual feature or stroke and the third is used to represent the temporal position of each stroke in the sequence. Lastly, a Transformer encoder $C_e$ process $c_{in}$ producing the context output $c \in \mathbb{R}^{L \times \fdim}$. 

\paragraph{VAE Encoder and Decoder} We use a VAE conditioned on the context information $c$ to produce a reconstruction $\hat{s}_t$ of the target strokes.
We model the VAE encoder $E$ as a transformer decoder receiving as query input a \new{the target strokes $s_t \in \mathbb{R}^{k\times8}$, which are projected to the hidden dimension of the transformer $\fdim$ with a linear layer.} The input sequence is enriched with 3D sinusoidal positional encodings (\cite{vaswani2017attention}), and two learnable tokens corresponding to the output mean and variance of the posterior gaussian distribution $\mu_z, \sigma_z^2 = E(s_t,c)$. We make use of the learnable tokens as a way to obtain outputs from the transformer representing the input sequence pooled over the temporal dimension (\cite{petrovich21actor}). The context information is used to condition $E$ and is provided as key and value inputs, conditioning the encoder through cross-attention.

The VAE decoder $D$ produces the sequence of reconstructed target strokes $\hat{s}_t=D(z,c)$. Preliminary experiments show that directly predicting $\hat{s}_t$ from $z$ and the context $c$ is difficult, so we propose a decoder composed of two stages $D=D_2 \circ D_1$, each modeled as a separate transformer decoder. The initial decoder $D_1$ receives as query input 1D sinusoidal positional encodings (\cite{vaswani2017attention}) providing temporal information regarding target strokes and predicts stroke positions $\hat{x}_t=D_1(z,c)$ of the target strokes. The transformer decoder is conditioned with the cross-attention mechanism on the context $c$ and on the latent variable $z \sim \mathcal{N}(\mu_z,\sigma_z^2)$ which are received as key and value inputs. The second transformer decoder $D_2$ is responsible for inferring the remaining stroke parameters conditioned by $\hat{x}_t$. In order to provide precise information about the reference image in the neighborhood of each predicted position, we extract image features $f_x$ from $f$ corresponding to the predicted position of each stroke using bilinear sampling, \textit{i.e.} $f_{\hat{x}}\!=\!\texttt{bilinear}(f,\hat{x}_t)$. Similarly to the VAE encoder, the sampled features are enriched with 3D sinusoidal positional encodings and are used as query inputs to infer the remaining stroke parameters $(\hat{\rho}_t,\hat{\sigma}_t,\hat{\omega}_t)=D_2(z,c,f_{\hat{x}})$. As with $D_1$, $c$ and $z$ condition the decoder as key and value inputs in the cross-attention operation. Finally, the outputs of $D_1$ and $D_2$ are combined to form the reconstructed target strokes $\hat{s}_t = (\hat{x}_t,\hat{\rho}_t,\hat{\sigma}_t,\hat{\omega}_t)$.

\subsection{Training}
\label{sec:training}

We train our model using the $\beta$-VAE (\cite{higgins2017betavae}) objective with an isotropic Gaussian prior as the main driving loss:
\begin{equation}
\begin{split}
    \mathcal{L}_{\beta\mhyphen\mathrm{VAE}} &= \mathbb{E}_{z\sim E(s_t,c)} \left\|s_t - \hat{s}_t\right\|_2^2 \\ 
    & + \lambda_{\mathrm{KL}} \mathcal{D}_\mathrm{KL}(\mathcal{N}(\mu_z, \sigma_z^2) \| \mathcal{N}(0, 1)).
\end{split}
\end{equation}

In addition, we notice that imprecisions in the reconstruction of the stroke positions $\hat{x}_t$ may bring to a situation where the reconstructed stroke color $\hat{\rho}_t$ differs from the color of $I_\mathrm{ref}$ at $\hat{x}_t$ which we call $\tilde{\rho}_t = I_\mathrm{ref}(\hat{x}_t)$. This mismatch is caused by the model ignoring the reference image and predicting target colors by attending only to context strokes and latent code and leads to performance degradation. For this reason, we introduce a color reconstruction loss that fosters the model to produce output strokes whose color is coherent with $I_\mathrm{ref}$:
\begin{equation}
    \mathcal{L}_{\mathrm{col}} = \left\|\tilde{\rho} - \hat{\rho}\right\|_2^2.
\end{equation}

Moreover, we propose two additional regularization losses that are aimed at improving the visual results at inference time when the latent codes $z$ are sampled from the prior distribution rather than the posterior. First, we impose the same color reconstruction loss on the predicted strokes to improve color coherency: 
\begin{equation}
    \mathcal{L}_{\mathrm{col}}^{reg} = \mathbb{E}_{z\sim \mathcal{N}(0,1)} \left\|I_{\mathrm{ref}}(x(D(z,c))) - \rho(D(z,c))\right\|_2^2,
\end{equation}
where $x(\cdot)$ and $\rho(\cdot)$ represent function extracting respectively position and color from the tuple of stroke parameters. Second, we impose a distribution matching objective aimed at maximizing the similarity between the characteristics of predicted and dataset stroke sequences. In particular, we propose to explicitly maximize the likelihood of sampling from the prior a sequence of strokes that is compatible with the dataset stroke distribution. For each sequence of corresponding context and target strokes, we concatenate them forming vector $s=s_c\mathbin\Vert s_t=s^{t-k+1:t+k}$, and build the corresponding feature vector $\psi$ capturing the relations between neighboring strokes. The feature vector $\psi$ is computed by taking the concatenation of the stroke features computed as follows: 
\begin{equation}
\label{eq:stroke_features}
    \psi = {\mathbin\Vert}_{l=1}^{l_\mathrm{max}}\left( {\mathbin\Vert}_{i=1}^{L-l} (s^{i+l} - s^{i})\right),
\end{equation}
where $L=2k$ represents the sequence length and $l_\mathrm{max}$ represents the maximum distance between strokes for which to extract features. In the following, we denote as $\psi$ the features produced on dataset stroke sequences and as $\hat{\psi}$ the features produced on inferred stroke sequences.
To make the computation tractable and easy to optimize, we assume independence of each dimension and fit two multivariate gaussian distributions $\mathcal{N}(\mu_\psi, \Sigma_\psi)$ and $\mathcal{N}(\mu_{\hat{\psi}}, \Sigma_{\hat{\psi}})$, respectively on $\psi$ and $\hat{\psi}$. Successively, we minimize KL divergence between the dataset distribution and the generated strokes distribution:
\begin{equation}
    \mathcal{L}_{\mathrm{dist}}^{reg} = \mathbb{E}_{z\sim \mathcal{N}(0,1)} \mathcal{D}_\mathrm{KL}(\mathcal{N}(\mu_{\hat{\psi}}, \Sigma_{\hat{\psi}}) \| \mathcal{N}(\mu_\psi, \Sigma_\psi)).
\end{equation}
The final optimization objective is given by:
\begin{equation}
\mathcal{L} = \mathcal{L}_{\beta\mhyphen\mathrm{VAE}} + \lambda_{\mathrm{col}} \mathcal{L}_{\mathrm{col}} + \lambda_{\mathrm{col}}^{reg} \mathcal{L}_{\mathrm{col}}^{reg} + \lambda_{\mathrm{dist}}^{reg} \mathcal{L}_{\mathrm{dist}}^{reg},
\label{eq:fianl_objective}
\end{equation}
where $\lambda_{\mathrm{col}}$, $\lambda_{\mathrm{col}}^{reg}$ and $\lambda_{\mathrm{dist}}^{reg}$ represent positive weighting terms. We show details of our training procedure in the \emph{Supp. Mat.}.

\subsection{Inference}
\label{sec:inference}

At inference time, our model is used iteratively to assist users in the creation of paintings corresponding to a reference image $I_\mathrm{ref}$. We consider the current state of the canvas as $I_c$ and the last $k$ strokes drawn on the canvas as the context strokes $s_c$. The context encoder $C$ is used to extract the context representation $c$. We note that, given the interactive scenario, the context strokes can originate either from the user or by a previous iteration of the model.
We then sample a latent vector $z$ from the prior distribution $\mathcal{N}(0, 1)$ and use the decoder to produce a plausible continuation of the painting process $D(z,c)$. By repeatedly sampling the latent vector $z$ from the prior distribution and keeping the context fixed, we provide a diverse set of plausible continuations of the painting from which the user can select the best option or keep drawing strokes manually if not satisfied by the proposals. The process is iterated until the painting is complete.

\subsection{Implementation Details}
\label{sec:implementation_details}
Following \cite{liu2021painttransformer}, we set the sequence length $k\!=\!8$ in all our experiments. We observe that, at inference time, the effective value of $k$ can be smaller, with the user selecting a sub-sequence of the proposed continuation. \new{Likewise, the length of the context strokes $s_c$ is set to $k=8$, which can be modified at inference time according to the user needs. We set the image resolution of $I_{\mathrm{ref}}, I_c$ to $H\!=\!W\!=\!256$, and implement the context feature extractor $F$ as a convolutional network reducing the spatial dimension of the input by a factor of 8, thus resulting in a feature map $\fvisual$ of size $H'\!=\!W'\!=16$. This makes the effective length of  $c_{in}$, the input of the context encoder, $L\!=\!256 + 8$. We model the context encoder $C_e$ as a transformer encoder, while the VAE encoder $E$ and VAE decoder $D$ are implemented as transformer decoders.  In all the cases, we set the hidden dimension of the models to $\fdim\!=\!256$.}
We train the final model for 5000 epochs, with a batch size of 32, using the AdamW optimizer (\cite{LoshchilovH19}) with initial learning rate of $1\mathrm{e}{-4}$ and cosine scheduler. We select the weights of each loss component in Eq.~\eqref{eq:fianl_objective} with a grid-search on the \textit{Oxford-IIIT Pet INP}, and apply the same configuration for experiments on the \textit{ADE 20K Outdoor INP}. The weight of each loss component is, respectively, $\lambda_{\mathrm{KL}}\!=\!2.5\mathrm{e}{-4}$, $\lambda_{\mathrm{col}}\!=\!2.5\mathrm{e}{-2}$,
$\lambda^{reg}_{\mathrm{col}}\!=\!2.5\mathrm{e}{-3}$ and $\lambda^{reg}_{\mathrm{dist}}\!=\!5.0\mathrm{e}{-6}$.
Additional implementation details are present in the \emph{Supp. Mat.}.

%% file: paper_sections/main_experiments.tex
\section{Experiments}
\label{sec_experiments}

In this section, we perform an experimental evaluation of the proposed method for INP. Sec.~\ref{sec:dataset} describes the adopted datasets, Sec.~\ref{sec:metrics} describes the adopted metrics, Sec.~\ref{sec:ablation} shows ablation results on our method, Sec.~\ref{sec:baselines} performs a quantitative comparison against baselines and Sec.~\ref{sec:qualitatives} shows qualitative results.

\subsection{Datasets}
\label{sec:dataset}

To train our architecture, we assume a dataset of images with an associated sequence of stroke parameters, representing the painting process used to produce the corresponding painting. To produce realistic stroke suggestions, our model captures the characteristics of the painting process represented in the dataset.
To overcome the cost associated with collecting human painting demonstrations, \elia{we follow recent work of \cite{cheng2022adaptively, singh2022paint2pix} and} choose to demonstrate that our framework is capable of modeling a painting process considering a synthetic dataset of stroke sequences that mimic a human painting process. Importantly, our method is general and learns the characteristics of the strokes provided as a demonstration, thus can be readily applied to a human-collected dataset if available.

We consider two existing image datasets and associate a sequence of strokes to each image, producing our INP datasets:

\begin{itemize}
    \item \textit{ADE 20K Outdoor INP}: we employ a subset of 5000 images of the ADE 20K dataset (\cite{zhou2017ade20k}) consisting of the set of original images depicting outdoor scenes. We split the dataset into a set of 4750 training images and 250 images for evaluation. 
    \item \textit{Oxford-IIIT Pet INP}: the dataset consists of 7349 images depicting different cat and dog breeds from \textit{Oxford-IIIT Pet} (\cite{parkhi2012cats}), both in indoor and outdoor scenarios. The dataset is split into 6980 training images and 369 images for evaluation.

\end{itemize}

Each image is decomposed into a sequence of strokes, parameterized as Sec.~\ref{sec:method}, using the NP method of \cite{zou2021stylized}. \new{Similar to the human painting process, the obtained sequence is organized in different levels of detail, with large strokes depicting the outline of the image first and fine-grained detail later in the sequence.} \st{where each stroke $s$ is translated to the RGB space using a pretrained differentiable renderer.} However, such sequences of strokes do not contain the sequential patterns typically produced by humans. Painters, in fact, due to the constraints imposed by physical brushes which discourage changes in color and brush, tend to produce sequences of strokes where the same color and brush sizes are maintained across several subsequent strokes. In addition, it is common for humans to produce paintings on an object-by-object basis (\cite{Singh2021IntelliPaintTD,zhao2020painting}) and to produce strokes in contiguous regions. To replicate these patterns \new{in our synthetic dataset and produce sequences with characteristics closer to real ones}, we perform a reordering of the stroke sequence produced by \cite{zou2021stylized} by optimizing a cost function that penalizes sequences with large differences in size, position or color between adjacent strokes or where adjacent strokes are placed on different subjects. Specifically, we  perform a reordering of the sequences by minimizing the following cost function, computed along the complete stroke sequence:
 
 \begin{align}
\mathrm{cost} &= \sum_{t=2}^{T} (\lambda^{\mathrm{ord}}_x \left\|x^{t} - x^{t-1}\right\|_2^2 + \lambda^{\mathrm{ord}}_\rho \left\|\rho^{t} - \rho^{t-1}\right\|_2^2\notag\\ &+ \lambda^{\mathrm{ord}}_\sigma(\sigma^{t} - \sigma^{t-1})^2 + \lambda_{\mathrm{obj}} \chi(x^t, x^{t-1}))
\end{align}

where $\lambda^{\mathrm{ord}}_x$, $\lambda^{\mathrm{ord}}_\rho$, $\lambda^{\mathrm{ord}}_\sigma$, $\lambda_{\mathrm{obj}}$ are positive weighting parameters. The function $\chi(x^t, x^{t-1})$ is equal to 1 if the input strokes are located on different subjects and 0 otherwise, and it is computed using the dataset segmentation masks.
\st{Importantly,}We ensure that the ordering relation between overlapping strokes is preserved, guaranteeing that both the original and reordered sequences of strokes produce the same visual output when rendered. Such a problem is an instance of the Sequential Ordering Problem which we optimize following \cite{helsgaun2017lkh3}. 
\input{tables/ablation_losses_complete}
\input{tables/ablation_architecture_complete}
\input{tables/baselines}

\subsection{Evaluation Metrics}
\label{sec:metrics}
\new{As outlined in Sec.~\ref{sec:problem_formulation}, the key three factors that we expect in the INP setting are:}
\st{To evaluate the effectiveness of our approach on the novel INP task, i.e. if it proposes diverse and plausible stroke continuations, we devise a set of metrics that capture the following key factors:} (i) the method is painting the reference image, (ii) the produced strokes parameters have characteristics that match the ones of the stroke dataset, (iii) diverse stroke continuations can be produced for the same context. We devise a set of quantitative metrics to capture these desiderata, and describe them in the following:
\st{In practice, we evaluate the method with the following quantitative metrics:}

\begin{itemize}
    \item Stroke Color L2 (L2) (i): we measure the L2 difference between the color of the predicted strokes and the color of the underlying reference image region. To avoid big strokes from dominating the metric, the L2 distance associated with each stroke is normalized by the stroke area before averaging. 
    \item Fr\'echet Stroke Distance (FSD) (ii): inspired by FID \cite{heusel2017advances}, we introduce a metric measuring the similarity between ground truth and predicted stroke sequences. For each sequence of context and target strokes, we compute stroke features $\psi$ as in Eq. \eqref{eq:stroke_features}. We report the Frechet Distance (\cite{frechet1957distance}) between the distribution of features derived from ground truth sequences and predicted ones.
    \item Fr\'echet Video Distance (FVD) (\cite{unterthiner2018towards}) (ii): given a sequence of context and target strokes, we generate videos of the corresponding canvas rendered up to each stroke in the sequence. We use FVD between videos produced with ground truth and predicted strokes as a metric for capturing the similarity between sequences.
    \item Wasserstein Distance (WD) (\cite{lantorovich1939wasserstein}) (ii): following \cite{liu2021painttransformer}, we adopt the Wasserstein Distance between Gaussian distributions fitted on the ground truth and predicted strokes as a stroke reconstruction quality metric. 
    \item Dynamic Time Warping (DTW) (\cite{muller2007dynamictimewarping}) (ii): we employ DTW between the ground truth and inferred target strokes to measure the quality of matching.
    \item LPIPS (\cite{zhang2018unreasonable}) (iii): following \cite{zhu2017toward}, we use LPIPS as a metric to compute the diversity of the produced outputs. For each reference image and context, we produce 5 stroke predictions and measure the average LPIPS diversity between all pairs of rendered results.
\end{itemize}

For each image in the test set, we extract 5 sequences of corresponding context and target strokes and compute the metrics on these samples. We note that WD and DTW require paired sequences of ground truth and reconstructed target sequences, while at inference time our method generates plausible stroke sequences that may not match the ground truth. \willi{For these metrics, we adopt a top-1 sampling strategy (\cite{yu2021diverse}) and generate 100 plausible stroke sequences, reporting the metric obtained for the best one.}

\subsection{Ablation Study}
\label{sec:ablation}
In this section, we ablate the main losses and architectural components of the proposed method. To improve the number of analyzed model configurations, in this section we reduce the number of training epochs to 1000.  
We start our analysis by ablating the contribution of the proposed losses (see Tab.~\ref{tab:ablation_losses}). 
Training the model only with the $\beta\mhyphen$VAE loss produces stroke outputs with high diversity but whose Stroke Color L2 is the highest in all the configurations, suggesting that the model is predicting strokes that are not consistent with the reference image.
Introducing $\mathcal{L}_{\mathrm{col}}$ promotes the model to take $I_\mathrm{ref}$ into account, resulting in a consistent reduction of the Stroke Color L2.
To further improve the performance, we introduce our two training regularization losses, which are aimed at improving quality when the latent code $z$ is sampled from the prior distribution at inference time.
Introducing $\mathcal{L}^{reg}_{\mathrm{col}}$ improves color accuracy as demonstrated by the best Stroke Color L2, but prevents the model from learning the users' painting style resulting in the highest FSD. 
Vice-versa, with only our proposed $\mathcal{L}_{\mathrm{dist}}^{reg}$ we can effectively learn the distribution of strokes, achieving the best performances in terms of FSD, WD, and LPIPS, but performance decreases in terms of the Stroke Color L2.
Only when combining all the proposed losses in our full model we obtain good performance under \textit{all} metrics. \\

Next, to ablate the contribution of each proposed architectural component, we produce the following modified versions of our method: (i) remove the context information provided by $s_c$ and $I_c$, the only context information comes from the reference image $I_{\mathrm{ref}}$; (ii) remove $C_e$; (iii) remove the two-step decoding procedure and replace it with a single transformer decoder that directly predicts $\hat{s}_t$; 
We show the ablation results in Tab.~\ref{tab:ablation_architecture}.
As expected, removing $s_c$ and $I_c$ from the context information significantly degrades the performance. Interestingly, LPIPS is the highest among the configurations, probably because, without conditioning from the context, the predictions can vary more freely. Likewise, removing the transformer encoder block $C_e$ consistently reduces the metrics, showing the importance of the module that provides richer context information to the decoder $D$ by combining visual and strokes features.
Configuration (iii) shows the impact of the two-step decoding procedure, designed to provide the model with richer visual information. This version of the method results in a degraded Stroke Color L2, FSD and FVD, showing the importance of detailed visual features in the prediction of strokes. 

\subsection{Comparison against Baselines}
\label{sec:baselines}
In this section, we compare our method against the state-of-the-art NP methods of \cite{zou2021stylized} and of \cite{liu2021painttransformer}. Due to the novelty of the task, these works are not directly comparable with the proposed one since they do not consider interaction, and need to be adapted to the INP setting.
A key component of the selected methods that makes them unsuitable for INP is their hierarchical rendering pipeline that iteratively divides the reference image and the canvas into smaller regions and operates on each in separation. This procedure allows the models to progressively focus on finer details and accurately reconstruct the reference image, but produces poor performance in INP since at each iteration the method may be forced to output strokes in a region far from the area the user is painting.
To avoid this limitation and make the model aware of the context, instead of operating on hardcoded regions, at each iteration we consider as the current region a portion of the image centered on the last context stroke. We call such a region the context region.
In this way, we encourage the method to output a sequence of strokes in the neighborhood of the context. In addition, we make the size of the region proportional to the area of the context strokes $s_c$, fostering the models to produce strokes with a level of detail compatible with the one currently adopted by the user.
We apply this modification to both methods and produce the following baselines: 

\begin{itemize}
    \item Paint Transformer (PT) \cite{liu2021painttransformer}: the model makes a prediction in the context region, but no explicit knowledge about the distribution of the dataset stroke sequences can be leveraged.
    \item Stylized Neural Painting (SNP) \cite{zou2021stylized}: we randomly initialize the sequence of predicted strokes to lie in the context region and optimize their parameters with the original SNP objective. Note that the method is not expressly conditioned on the context strokes and does not consider the characteristics of the dataset stroke sequences.
    \item Stylized Neural Painting+ (SNP+) \cite{zou2021stylized}: we modify SNP to explicitly take into consideration the characteristics of the dataset stroke sequence. In detail, we modify the SNP optimization objective by introducing a term similar to the distribution matching loss $\mathcal{L}_{\mathrm{dist}}^{\mathrm{reg}}$ to produce stroke sequences whose features $\hat{\psi}$ match those of the training dataset $\psi$. We extract stroke features from the dataset using Eq. \eqref{eq:stroke_features} and fit a multivariate Gaussian distribution with independent components on $\psi$. We improve the realism of inferred strokes by maximizing the likelihood that the inferred features $\hat{\psi}$ match the fitted distribution.
\end{itemize}

Tab.~\ref{tab:baseline} shows comparison results of our method against the baselines. To ensure a fair comparison, we follow \cite{liu2021painttransformer} and set the length of the predicted sequence to $k\!=\!8$ for all the methods.
While baseline methods are designed with the main objective of producing strokes that closely match the reference image, we notice that our method presents a Stroke Color L2 metric similar to SNP and SNP+, and lower with respect to PT.
Moreover, Paint Transformer and Stylized Neural Painting have, by design, no way to leverage information about the characteristics of the dataset strokes and tend to produce stroke sequences whose characteristics do not match the ones in the dataset. This is highly reflected in the metrics that capture the ability to paint like the demonstration; our method strongly outperforms PT and SNP in terms of the WD, DTW, FVD, and FSD. On the other side, SNP+ can exploit such information. As expected, this greatly reduces the FSD compared to the naive SNP, but comes at the cost of increasing the Stroke Color L2. Interestingly, our method can outperform SNP+ in this metric, suggesting that the better performance of our model is due not only to distribution matching objectives but also to the architecture design.
Finally, we evaluate the capacity of the method to produce varied plausible outputs for a fixed context. We observe that, while PT is deterministic and no variability can be produced, diverse predictions for a given context can be obtained for SNP and SNP+ by starting the optimization from different randomly initialized stroke parameters. Our method instead is probabilistic by nature, with a conditional VAE designed to generate different plausible continuations for a fixed context. Despite the non-determinism of some baselines, the diversity of their predictions is inferior to the ones obtained with our method, which achieves the highest LPIPS diversity score.  

\paragraph{\new{User Study}} We complement our quantitative results with a user study. We show the users a set of reference videos with rendered stroke sequences from the dataset followed by two videos, one with rendered strokes produced by one of the baselines, and one produced by \methodname. \new{To ease the evaluation, we produce sequences with a length of
24 strokes and render the obtained strokes in a short video.} We ask the users to express which of the two videos has strokes whose characteristics resemble the reference videos the most. We gather a total of 960 votes from 8 unique users. \new{We report the results in Tab.~\ref{tab:user_study},} \st{Users express a preference of 97.9\%, 97.1\%, and 95.0\% for our method respectively against PT, SNP, and SNP+,} showing a clear preference for \methodname~when evaluated on the INP task (see \emph{Supp. Mat.} for additional details).
\input{tables/user_study}

\subsection{Qualitative Results}

\input{tables/qualitative_baselines}
\input{tables/qualitative_heatmap}
\input{tables/qualitative_diversity}
\input{tables/qualitative_interpolation}

We provide qualitative results comparing our method with the baselines. \elia{Visualizing results as still images provide limited understanding of the INP task, we refer the reader to \emph{Supp.~Mat.}~for video results along with a working demo. When comparing different methods, the predicted stroke sequences should satisfy three criteria: (i) make the canvas more similar to the reference image, (ii) present similar characteristics as the demonstrations, (iii) provide diverse continuations (Sec.~\ref{sec:problem_formulation}, \ref{sec:metrics})}. \elia{While all the methods perform similarly when assessing (i), \methodname~strongly outperform
the competitors according to (ii) and (iii) which are peculiar to the \textit{interactive} component of the task.} \\

First, we analyze the ability of \methodname~to model the characteristics of the dataset strokes \elia{(ii)}. Given the reference image and the context fixed, we sample 100 possible stroke continuations. A method that correctly captures the original stroke distribution is expected to yield at least a stroke continuation close to the ground truth. In Fig.~\ref{fig:qualitative_baseline} we plot qualitative results showing the sampled sequence that better matches the ground truth in terms of L2 distance. In all examples, our method is able to produce a sequence close to the ground truth, while PT, SNP and SNP+ struggle to generate a matching sequence, indicating that our method is able to better capture the characteristics of the dataset stroke sequences. 
\elia{In Fig.~\ref{fig:qualitative_heatmap} we show the heatmap depicting the probability, obtained with the aforementioned sampling procedure, that a certain pixel will be covered by one of the next $k$ strokes. 
We notice that only \methodname~is able to produce different plausible suggestions (iii) while predicting strokes on the same object as the last context strokes (ii)}.
Second, we further evaluate the ability of \methodname~to predict different stroke continuations given a fixed context \elia{(iii)}.
Note that, as shown in Fig.~\ref{fig:teaser} (b), this is an important feature of an INP method since at each iteration of the painting process the method should be able to propose at least a stroke sequence that matches the user painting intentions.
In Fig.~\ref{fig:qualitative_diversity} we show different stroke suggestions for a fixed context obtained by sampling different latent codes $z$ from the unit normal prior distribution $\mathcal{N}(0, 1)$. Our method is capable of generating diverse stroke continuations, each of which focuses on a similar region, color and subject with respect to the given context. \elia{Finally, we show the structure of the learned latent representation (Fig. \ref{fig:interpolation}). Specifically, we sample two different latent codes from the prior distribution, $z_{\mathrm{start}}, z_{\mathrm{end}} \sim \mathcal{N}(0,1)$ and we linearly interpolate between the two samples, plotting the predicted results along the interpolation path (see \emph{Supp. Mat.} for video animation). 
It is possible to notice that the strokes smoothly transition between the two samples $z_{\mathrm{start}}, z_{\mathrm{end}}$, changing their position but focusing on the same subject suggesting that the learned latent space is well structured.}

\label{sec:qualitatives}

%% file: tables/ablation_losses_complete.tex
\begin{table*}[t]
\centering
\setlength{\tabcolsep}{2.0pt}
\footnotesize
\begin{tabular}{lcccccccccccccccc}
\toprule
\multicolumn{1}{l}{} & \multicolumn{4}{c}{} & \multicolumn{6}{c}{\emph{ADE 20K Outdoor INP}} & \multicolumn{6}{c}{\emph{Oxford-IIIT Pet INP}} \\
\cmidrule(lr){6-11}
\cmidrule(lr){12-17}
 & $\mathcal{L}_{\beta\mhyphen\mathrm{VAE}}$ & $\mathcal{L}_{\mathrm{col}}$ & $\mathcal{L}_{\mathrm{col}}^{reg}$ & $\mathcal{L}_{\mathrm{dist}}^{reg}$ & L2$\downarrow$ & FSD$\downarrow$ & FVD$\downarrow$ & WD$\downarrow$ & DTW$\downarrow$ & LPIPS$\uparrow$ & L2$\downarrow$ & FSD$\downarrow$ & FVD$\downarrow$ & WD$\downarrow$ & DTW$\downarrow$ & LPIPS$\uparrow$ \\
\midrule
&\cmark &\xmark & \xmark & \xmark &  0.136	& \textbf{1.64} &  11.9 & \textbf{0.032} & \textbf{0.849} & 0.038 &  0.155 & 1.29 & 13.2 & \textbf{0.031} & \textbf{0.851} & 0.038 \\
&\cmark &\cmark & \xmark & \xmark & 0.058 & 2.44 & 7.18 & 0.034 & 0.899 & 0.031 & 0.057 & 2.05 & 7.31 & 0.033 & 0.910 &  0.029 \\
&\cmark &\cmark & \cmark & \xmark & \textbf{0.043} & 6.84 & 8.16 & 0.040 & 0.974	& 0.028 & \textbf{0.039} & 5.17 & 6.77 & 0.035 & 0.942 & 0.030 \\
&\cmark &\cmark & \xmark & \cmark & 0.094 & 1.92 & 10.5 & 0.033 & 0.892 & \textbf{0.044} & 0.091 & \textbf{1.16} & 9.51 & \textbf{0.031} & 0.893 & \textbf{0.039} \\
\midrule
Full &\cmark &\cmark&\cmark&\cmark& 0.044 & 2.04 & \textbf{6.60} & 0.034 & 0.893 & 0.033 & 0.042 &	1.51 &	\textbf{6.72} & 0.032 &	0.893 & 0.030 \\
\bottomrule
\end{tabular}
\caption{Loss ablation results on the \emph{ADE 20K Outdoor INP} dataset and the \textit{Oxford-IIIT Pet INP} dataset.}
\label{tab:ablation_losses}
\end{table*}

%% file: tables/ablation_architecture_complete.tex
\begin{table*}[t]
\centering
\setlength{\tabcolsep}{2.0pt}
\footnotesize
\begin{tabular}{lcccccccccccc}
\toprule
\multicolumn{1}{l}{} & \multicolumn{6}{c}{\emph{ADE 20K Outdoor INP}} & \multicolumn{6}{c}{\emph{Oxford-IIIT Pet INP}} \\
\cmidrule(lr){2-7}
\cmidrule(lr){8-13}

Model Version & L2$\downarrow$ & FSD$\downarrow$ & FVD$\downarrow$ & WD$\downarrow$ & DTW$\downarrow$ & LPIPS$\uparrow$ & L2$\downarrow$ & FSD$\downarrow$ & FVD$\downarrow$ & WD$\downarrow$ & DTW$\downarrow$ & LPIPS$\uparrow$\\
\midrule
(i) &  0.060 &	16.4 &	21.7 & 0.058 &	1.11 &	\textbf{0.039} &  0.059 &	11.9 & 23.3 & 0.058 & 1.13 & \textbf{0.037} \\
(ii) &  0.050 &	2.27 &	7.45 &	0.035 &	0.907 &	0.031 &  0.048 & 1.87 & 7.78 & 0.036 & 0.915 &  0.027 \\
(iii) &  0.052 & 2.28 & 7.12 & \textbf{0.032} & \textbf{0.868} &	0.029  &  0.049 &	1.70 & 7.83 & 0.033 &	\textbf{0.893} &	0.028\\
\midrule
Full &  \textbf{0.044} & \textbf{2.04} & \textbf{6.60} & 0.034 & 0.893 & 0.033 &  \textbf{0.042} & \textbf{1.51} & \textbf{6.72} & \textbf{0.032} & \textbf{0.893} & 0.030 \\
\bottomrule
\end{tabular}
\caption{Ablation architectural choices on \emph{ADE 20K Outdoor INP} and \textit{Oxford-IIIT Pet INP} datasets. Each row represent a different model version, respectively: (i) remove context information $I_C$ and $s_c$, (ii) remove $C_e$, (iii) decode all strokes parameters with a single-step decoder and our full model.
}
\label{tab:ablation_architecture}
\end{table*}

%% file: tables/baselines.tex
\begin{table*}[!ht]
\centering
\setlength{\tabcolsep}{2.0pt}
\footnotesize
\begin{tabular}{lcccccccccccc}
\toprule
\multicolumn{1}{c}{} & \multicolumn{6}{c}{\emph{ADE 20K Outdoor INP}} & \multicolumn{6}{c}{\emph{Oxford-IIIT Pet INP}} \\
\cmidrule(lr){2-7}
\cmidrule(lr){8-13}
Method & L2$\downarrow$ & FSD$\downarrow$ & FVD$\downarrow$ & WD$\downarrow$ & DTW$\downarrow$ & LPIPS$\uparrow$ & L2$\downarrow$ & FSD$\downarrow$ & FVD$\downarrow$ & WD$\downarrow$ & DTW$\downarrow$ & LPIPS$\uparrow$\\
\midrule
PT (\cite{liu2021painttransformer}) &  0.056 & 10.6 &	9.06 & 0.073 & 1.41	& 0	& 0.048	& 11.3	& 8.77	& 0.074	& 1.47	& 0 \\
SNP (\cite{zou2021stylized}) & 0.044	& 13.7 & 7.05 & 0.082 & 1.25	& 0.018	& \textbf{0.037}	& 16.3	& 6.09	& 0.075	& 1.27	& 0.017 \\
SNP+ (\cite{zou2021stylized}) & 0.045 &	8.50 & 7.20 & 0.081 & 1.16	& 0.017	& 0.039	& 9.57	& 5.95	& 0.074	& 1.20	& 0.017 \\
\midrule

\methodname & \textbf{0.040} &	\textbf{1.50} & \textbf{6.27} & \textbf{0.031} & \textbf{0.876}	& \textbf{0.032}	& \textbf{0.037}	& \textbf{1.12}	& \textbf{4.73}	& \textbf{0.029}	& \textbf{0.867}	& \textbf{0.031}  \\

\bottomrule
\end{tabular}
\caption{Comparison with baselines on \emph{ADE 20K Outdoor INP} and \emph{Oxford-IIIT Pet INP} datasets.}
\label{tab:baseline}
\end{table*}

%% file: tables/user_study.tex
\begin{table}[!t]
\begin{center}
\setlength{\tabcolsep}{1.0pt}
\begin{tabular}{lccc}
\toprule
 & PT & SNP & SNP+ \\
 & \cite{liu2021painttransformer} & \cite{zou2021stylized} & \cite{zou2021stylized} \\
\midrule
Preferences & 97.9\% & 97.1\% & 95.0\%  \\
\bottomrule
\end{tabular}
\end{center}
\caption{User study comparing the preferences between \methodname~and the respective baseline.}
\label{tab:user_study}
\end{table}

%% file: tables/qualitative_baselines.tex
\begin{table}[!bh]
    \centering
    \def\arraystretch{1.0}
    \resizebox{\columnwidth}{!}{
    \setlength\tabcolsep{0pt}
    \footnotesize
    \renewcommand{\arraystretch}{0.0}
    \begin{tabular}{ccccc}
         $I_{\text{ref}}$ & \methodname & PT & SNP & SNP+  \\
         
        \includegraphics[trim=2 0 0 2,clip,width=0.2\columnwidth]{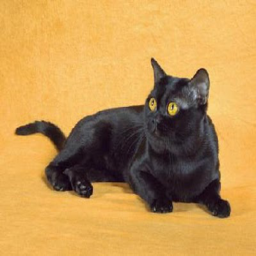} & 
        \includegraphics[trim=2 0 0 2,clip,width=0.2\columnwidth]{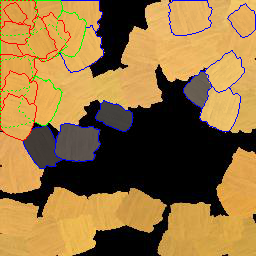} & 
        \includegraphics[trim=2 0 0 2,clip,width=0.2\columnwidth]{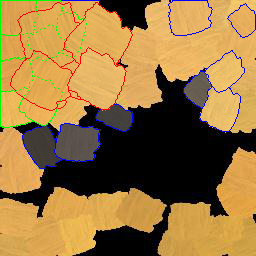} & 
        \includegraphics[trim=2 0 0 2,clip,width=0.2\columnwidth]{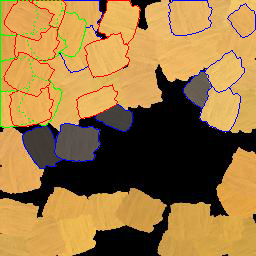} & 
        \includegraphics[trim=2 0 0 2,clip,width=0.2\columnwidth]{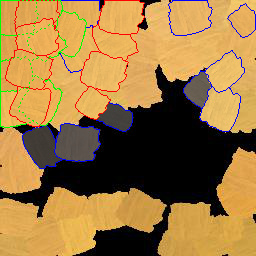} \\
        
                 \includegraphics[trim=2 0 0 2,clip,width=0.2\columnwidth]{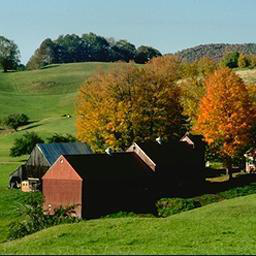} & 
         \includegraphics[trim=2 0 0 2,clip,width=0.2\columnwidth]{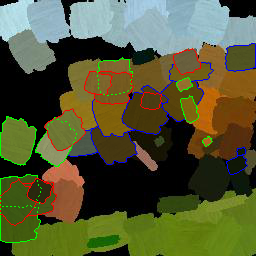} & 
         \includegraphics[trim=2 0 0 2,clip,width=0.2\columnwidth]{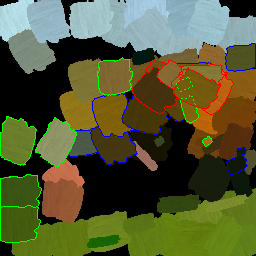} & 
         \includegraphics[trim=2 0 0 2,clip,width=0.2\columnwidth]{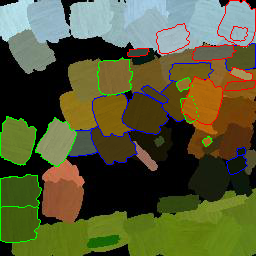} & 
         \includegraphics[trim=2 0 0 2,clip,width=0.2\columnwidth]{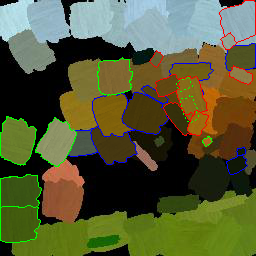} \\
         
         \includegraphics[trim=2 0 0 2,clip,width=0.2\columnwidth]{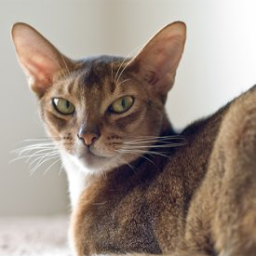} & 
         \includegraphics[trim=2 0 0 2,clip,width=0.2\columnwidth]{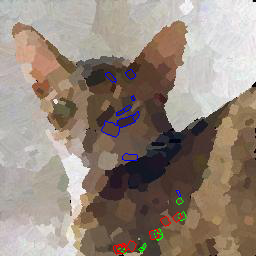} & 
         \includegraphics[trim=2 0 0 2,clip,width=0.2\columnwidth]{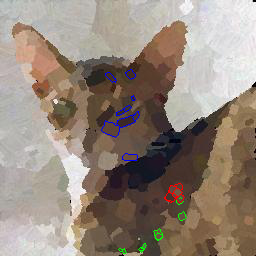} & 
         \includegraphics[trim=2 0 0 2,clip,width=0.2\columnwidth]{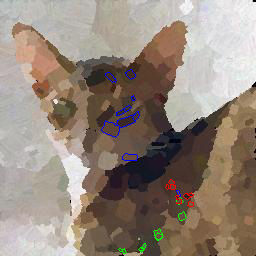} & 
         \includegraphics[trim=2 0 0 2,clip,width=0.2\columnwidth]{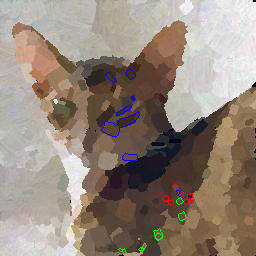} \\
         
        \includegraphics[trim=2 0 0 2,clip,width=0.2\columnwidth]{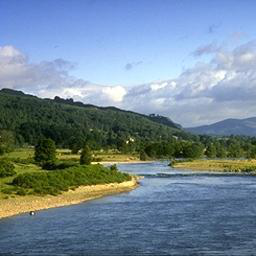} & 
        \includegraphics[trim=2 0 0 2,clip,width=0.2\columnwidth]{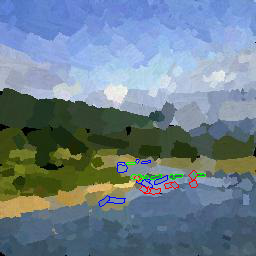} & 
        \includegraphics[trim=2 0 0 2,clip,width=0.2\columnwidth]{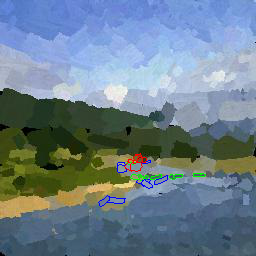} & 
        \includegraphics[trim=2 0 0 2,clip,width=0.2\columnwidth]{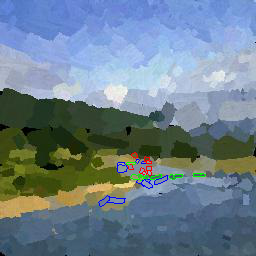} & 
        \includegraphics[trim=2 0 0 2,clip,width=0.2\columnwidth]{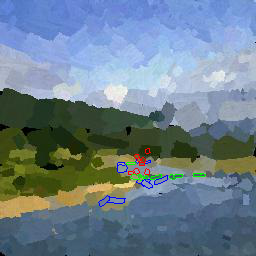} \\
    \end{tabular}
    }
    \vspace{2mm}
    \captionof{figure}{Qualitative comparison of \methodname~with baselines. Given the {\color{blue} context} (blue) strokes, we generate 100 {\color{red} predicted} (red) stroke sequences and plot the one that better matches the {\color{green} ground truth} (green). 
    \elia{First row, only \methodname~is able to produce a sequence whose stroke positioning is similar to the ones of the ground truth, while the baselines tend to unrealistically cluster the strokes in a tight area. Second row, successive strokes predicted by \methodname~have similar colors as in the dataset demonstrations, while the baselines unrealistically jump between the grass, the sky, and the trees}} 
    \label{fig:qualitative_baseline}
\end{table}

%% file: tables/qualitative_heatmap.tex
\begin{table}[hbt]
    \centering
    \def\arraystretch{1.0}
    \resizebox{\columnwidth}{!}{
    \setlength\tabcolsep{0pt}
    \footnotesize
    \renewcommand{\arraystretch}{0.0}
    \begin{tabular}{ccccc}
         $I_{\text{c}}$ & \methodname & PT & SNP & SNP+  \\
                        
            \includegraphics[trim=2 0 0 2,clip,width=0.2\columnwidth]{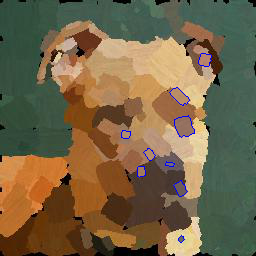} &
            \includegraphics[trim=2 0 0 2,clip,width=0.2\columnwidth]{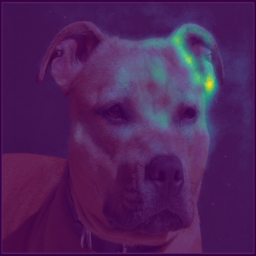} &
            \includegraphics[trim=2 0 0 2,clip,width=0.2\columnwidth]{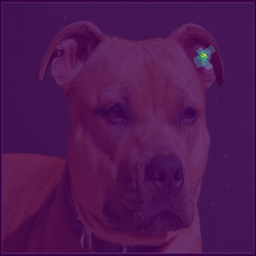} &
            \includegraphics[trim=2 0 0 2,clip,width=0.2\columnwidth]{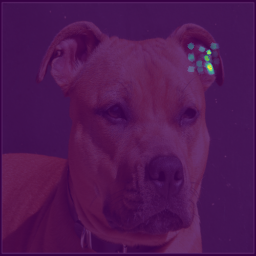} &
            \includegraphics[trim=2 0 0 2,clip,width=0.2\columnwidth]{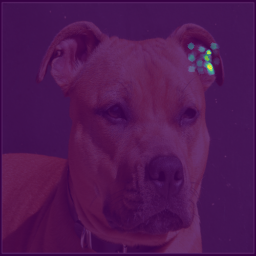} \\
            
            \includegraphics[trim=2 0 0 2,clip,width=0.2\columnwidth]{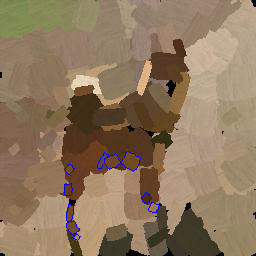} &
            \includegraphics[trim=2 0 0 2,clip,width=0.2\columnwidth]{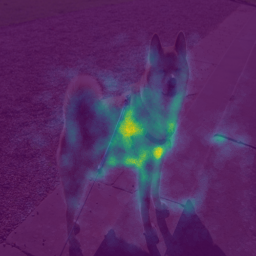} &
            \includegraphics[trim=2 0 0 2,clip,width=0.2\columnwidth]{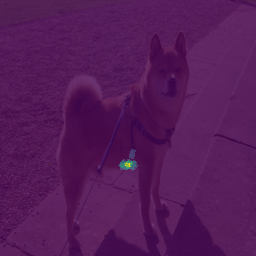} &
            \includegraphics[trim=2 0 0 2,clip,width=0.2\columnwidth]{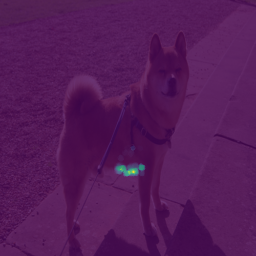} &
            \includegraphics[trim=2 0 0 2,clip,width=0.2\columnwidth]{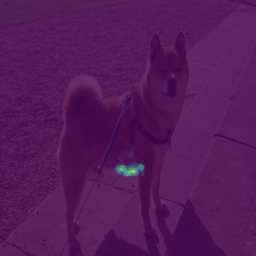} \\
            
            \includegraphics[trim=2 0 0 2,clip,width=0.2\columnwidth]{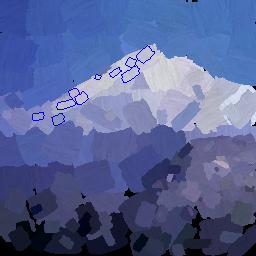} &
            \includegraphics[trim=2 0 0 2,clip,width=0.2\columnwidth]{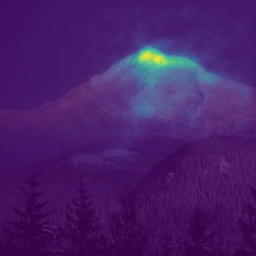} &
            \includegraphics[trim=2 0 0 2,clip,width=0.2\columnwidth]{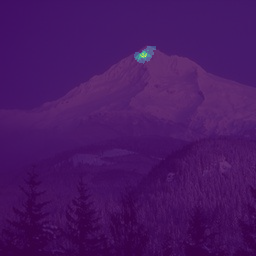} &
            \includegraphics[trim=2 0 0 2,clip,width=0.2\columnwidth]{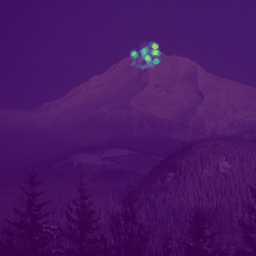} &
            \includegraphics[trim=2 0 0 2,clip,width=0.2\columnwidth]{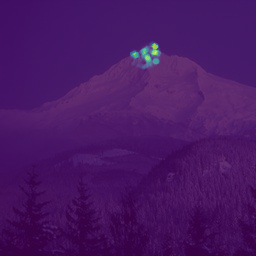}  \\
            
            \includegraphics[trim=2 0 0 2,clip,width=0.2\columnwidth]{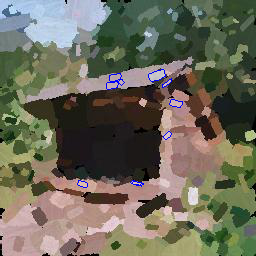} &
            \includegraphics[trim=2 0 0 2,clip,width=0.2\columnwidth]{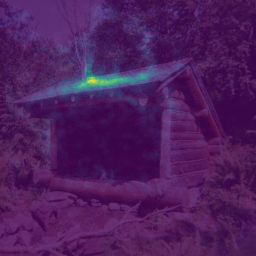} &
            \includegraphics[trim=2 0 0 2,clip,width=0.2\columnwidth]{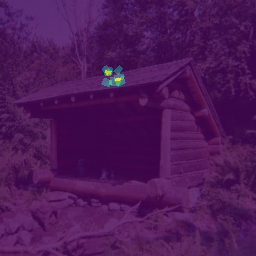} &
            \includegraphics[trim=2 0 0 2,clip,width=0.2\columnwidth]{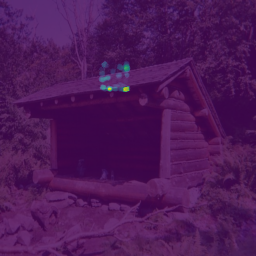} &
            \includegraphics[trim=2 0 0 2,clip,width=0.2\columnwidth]{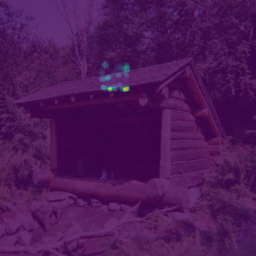}  \\
    \end{tabular}
    }
    \vspace{2mm}
    \captionof{figure}{Next predicted stroke probability distribution. For each method, we show the probability that a given pixel will be occupied by a predicted stroke for the given context. The context (outlined in blue in the first column) is kept fixed and we sample $n=500$ continuations for each method to estimate the probability distribution.}
    \label{fig:qualitative_heatmap}
\end{table}

%% file: tables/qualitative_diversity.tex
\begin{table}[!ht]
    \centering
    \def\arraystretch{1.0}
    \resizebox{\columnwidth}{!}{
    \setlength\tabcolsep{0pt}
    \footnotesize
    \renewcommand{\arraystretch}{0.0}
    \begin{tabular}{ccccc}
         $I_{\text{ref}}$ & $z_1$ & $z_2$ & $z_3$ & $z_4$  \\
         
        \includegraphics[trim=2 0 0 2,clip,width=0.2\columnwidth]{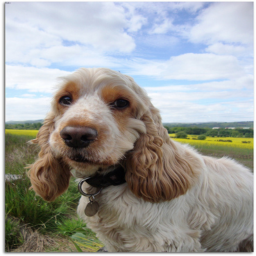} & 
        \includegraphics[trim=2 0 0 2,clip,width=0.2\columnwidth]{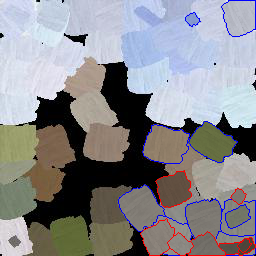} & 
        \includegraphics[trim=2 0 0 2,clip,width=0.2\columnwidth]{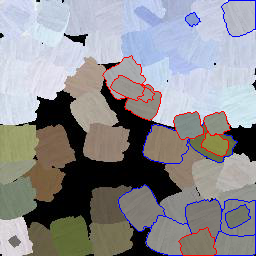} & 
        \includegraphics[trim=2 0 0 2,clip,width=0.2\columnwidth]{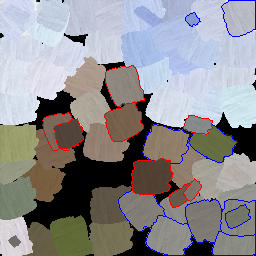} & 
        \includegraphics[trim=2 0 0 2,clip,width=0.2\columnwidth]{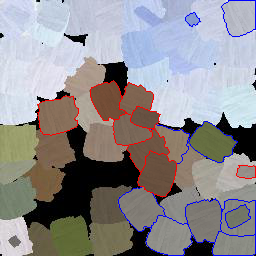} \\
		 
		 \includegraphics[trim=2 0 0 2,clip,width=0.2\columnwidth]{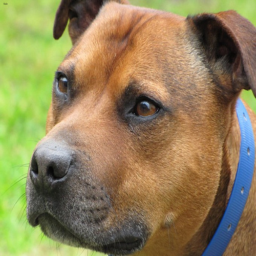} & 
         \includegraphics[trim=2 0 0 2,clip,width=0.2\columnwidth]{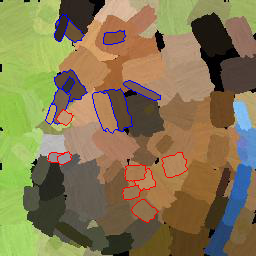} & 
         \includegraphics[trim=2 0 0 2,clip,width=0.2\columnwidth]{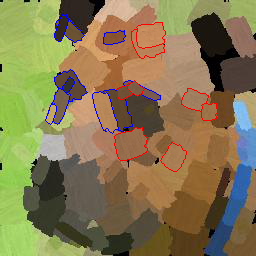} & 
         \includegraphics[trim=2 0 0 2,clip,width=0.2\columnwidth]{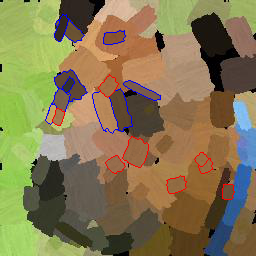} & 
         \includegraphics[trim=2 0 0 2,clip,width=0.2\columnwidth]{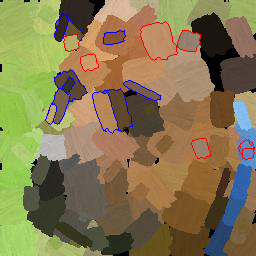} \\
         
 		 \includegraphics[trim=2 0 0 2,clip,width=0.2\columnwidth]{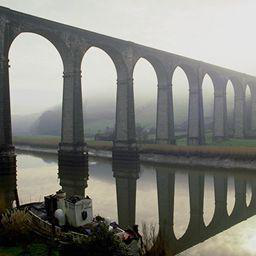} & 
         \includegraphics[trim=2 0 0 2,clip,width=0.2\columnwidth]{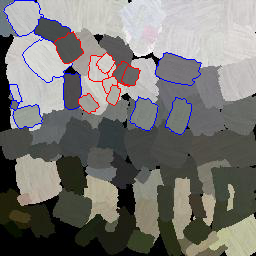} & 
         \includegraphics[trim=2 0 0 2,clip,width=0.2\columnwidth]{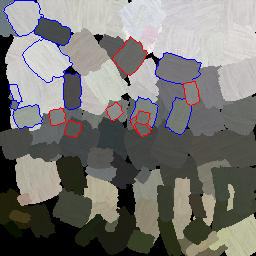} & 
         \includegraphics[trim=2 0 0 2,clip,width=0.2\columnwidth]{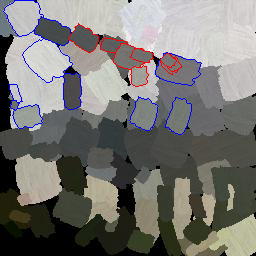} & 
         \includegraphics[trim=2 0 0 2,clip,width=0.2\columnwidth]{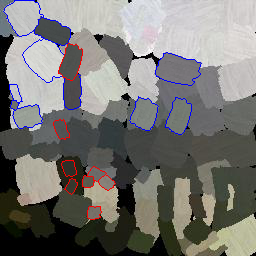} \\

 		 \includegraphics[trim=2 0 0 2,clip,width=0.2\columnwidth]{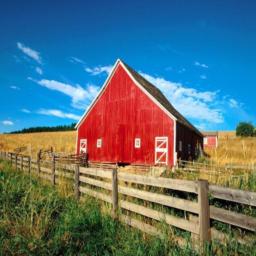} & 
          \includegraphics[trim=2 0 0 2,clip,width=0.2\columnwidth]{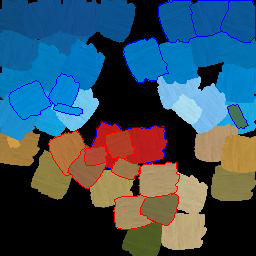} & 
          \includegraphics[trim=2 0 0 2,clip,width=0.2\columnwidth]{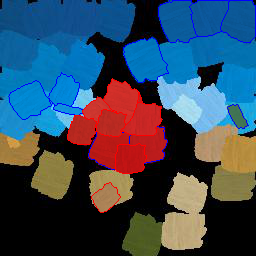} & 
          \includegraphics[trim=2 0 0 2,clip,width=0.2\columnwidth]{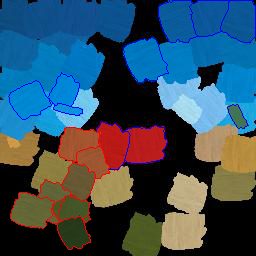} & 
          \includegraphics[trim=2 0 0 2,clip,width=0.2\columnwidth]{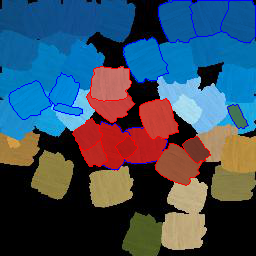} \\
         
    \end{tabular}
    }
    
    \vspace{2mm}
    \captionof{figure}{Diversity of proposed continuations. Given a fixed {\color{blue} context} (blue), we sample different $z_i \sim \mathcal{N}(0,1)$ and plot the correspondent {\color{red} predicted} (red) strokes in different columns.} 
    \label{fig:qualitative_diversity}
\end{table}

%% file: tables/qualitative_interpolation.tex
\begin{table}[!t]
    \centering
    \def\arraystretch{1.0}
    \resizebox{\columnwidth}{!}{
    \setlength\tabcolsep{0pt}
    \footnotesize
    \renewcommand{\arraystretch}{0.0}
    \begin{tabular}{cccccc}
         $\alpha\!=\!0$ & $\alpha\!=\!0.2$ & $\alpha\!=\!0.4$ & $\alpha\!=\!0.6$ & $\alpha\!=\!0.8$ & $\alpha\!=\!1$ \\
       \includegraphics[trim=2 0 0 2,clip,width=0.2\columnwidth]{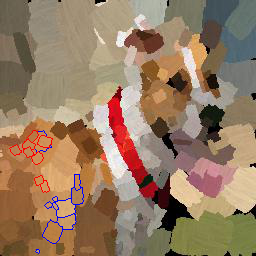} &
       \includegraphics[trim=2 0 0 2,clip,width=0.2\columnwidth]{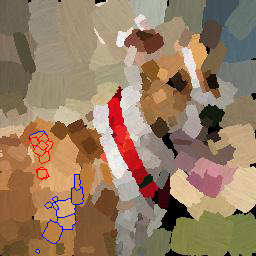} &
       \includegraphics[trim=2 0 0 2,clip,width=0.2\columnwidth]{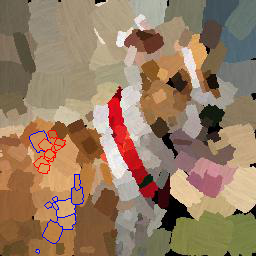} &
       \includegraphics[trim=2 0 0 2,clip,width=0.2\columnwidth]{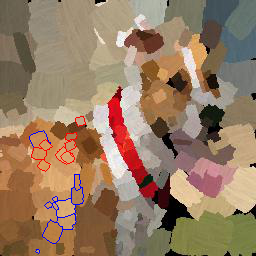} &
       \includegraphics[trim=2 0 0 2,clip,width=0.2\columnwidth]{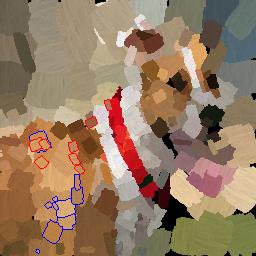} &
       \includegraphics[trim=2 0 0 2,clip,width=0.2\columnwidth]{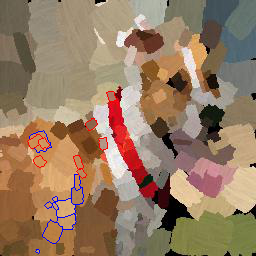} \\

       \includegraphics[trim=2 0 0 2,clip,width=0.2\columnwidth]{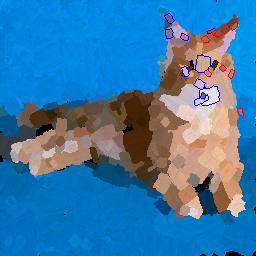} &
       \includegraphics[trim=2 0 0 2,clip,width=0.2\columnwidth]{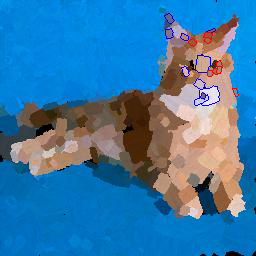} &
       \includegraphics[trim=2 0 0 2,clip,width=0.2\columnwidth]{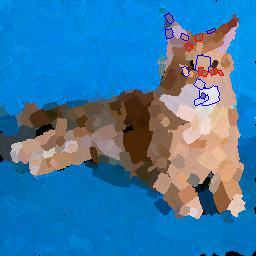} &
       \includegraphics[trim=2 0 0 2,clip,width=0.2\columnwidth]{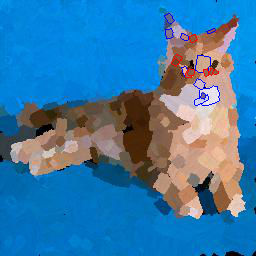} &
       \includegraphics[trim=2 0 0 2,clip,width=0.2\columnwidth]{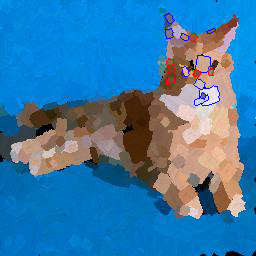} &
       \includegraphics[trim=2 0 0 2,clip,width=0.2\columnwidth]{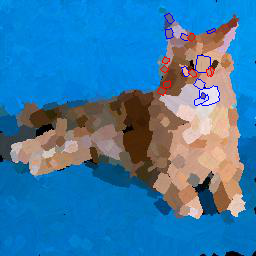} \\
       
        \includegraphics[trim=2 0 0 2,clip,width=0.2\columnwidth]{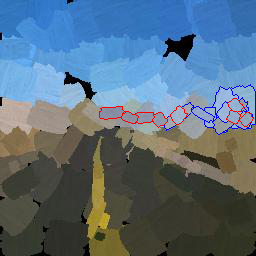} &
        \includegraphics[trim=2 0 0 2,clip,width=0.2\columnwidth]{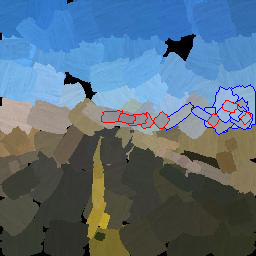} &
        \includegraphics[trim=2 0 0 2,clip,width=0.2\columnwidth]{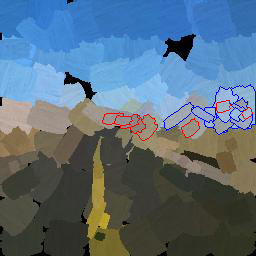} &
        \includegraphics[trim=2 0 0 2,clip,width=0.2\columnwidth]{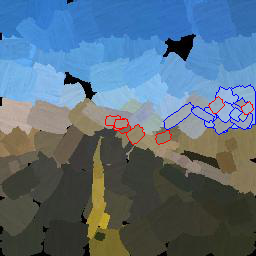} &
        \includegraphics[trim=2 0 0 2,clip,width=0.2\columnwidth]{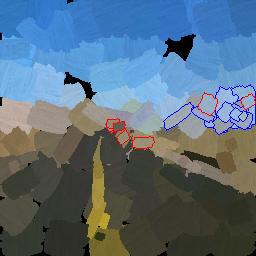} &
        \includegraphics[trim=2 0 0 2,clip,width=0.2\columnwidth]{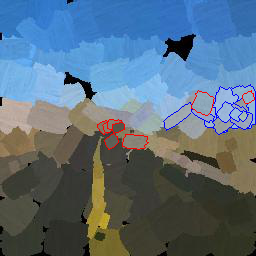} \\

        \includegraphics[trim=2 0 0 2,clip,width=0.2\columnwidth]{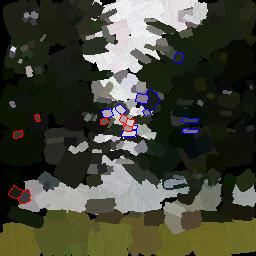} &
        \includegraphics[trim=2 0 0 2,clip,width=0.2\columnwidth]{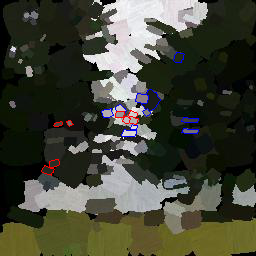} &
        \includegraphics[trim=2 0 0 2,clip,width=0.2\columnwidth]{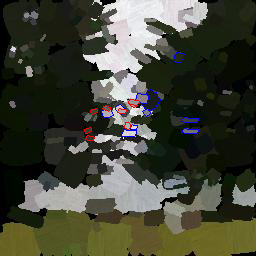} &
        \includegraphics[trim=2 0 0 2,clip,width=0.2\columnwidth]{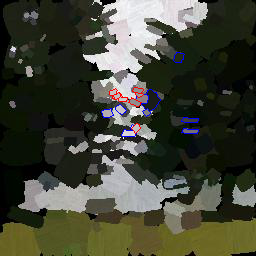} &
        \includegraphics[trim=2 0 0 2,clip,width=0.2\columnwidth]{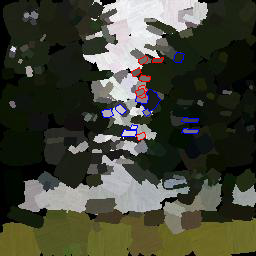} &
        \includegraphics[trim=2 0 0 2,clip,width=0.2\columnwidth]{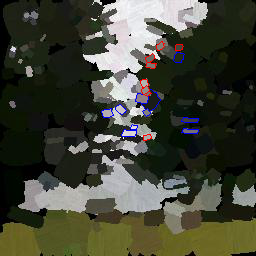}
    
    \end{tabular}
    }
    \vspace{2mm}
    \captionof{figure}{Latent code $z$ interpolation results (the reference image is omitted for better visualization). For each image, given the same {\color{blue} context} strokes, we sample two latent codes  $z_{\mathrm{start}}, z_{\mathrm{end}} \sim \mathcal{N}(0,1)$ and linearly interpolate, obtaining $z_i = (1 - \alpha) \cdot z_{\mathrm{start}} + \alpha \cdot z_{\mathrm{end}}$ (outlined in {\color{red} red}). The strokes smoothly transition changing position but focusing on the same object.}
    \label{fig:interpolation}
\end{table}

%% file: paper_sections/main_conclusions.tex
\section{Conclusions}
\label{sec_conclusion}

In this paper we introduce the novel task of interactive neural painting, where a user tries to reproduce a scene on a painting and the objective is to give (multiple) suggestions about the next strokes to paint, thus helping the user in producing its artwork. The proposed method is based on a conditional transformer VAE architecture, which is demonstrated on two novel datasets. Our experiments show that our model: correctly paints the reference image, outputs strokes whose characteristics are close to demonstration data, gives diverse yet plausible brush-stroke suggestions; and outperforms the analyzed baselines on a large set of metrics and on a user study. We hope that our work can stimulate further research in this domain.

%% file: paper_sections/appendix.tex
In this appendix, we provide additional details about parametrization and rendering of the strokes, a description of the architecture and the losses, and user study details. 

This appendix is complemented by a supplementary website where we show additional qualitative results in the form of videos. In particular, we showcase a demo of \methodname~in an Interactive Neural Painting (INP) scenario, where a user interacts with our model to paint a reference image. \new{The material is available at the \href{https://helia95.github.io/inp-website}{\textcolor{magenta}{project page}}.}

\subsection*{Dataset}
\label{sec:supp_dataset}
Due to the cost associated with the acquisition of a real stroke dataset, we evaluate our model on two synthetic stroke datasets, built to mimic human painting style (see Sec.~4.1 of the main paper for a discussion). Here we detail the dataset acquisition procedure. Examples from the dataset are provided on the supplementary website. \\

\noindent \emph{Images.} We rely on two publicly available datasets, each containing images and associated segmentation masks.
The \textit{Oxford-IIIT Pet} dataset (\cite{parkhi2012cats}), contains 7349 images of cats and dogs of different species for a total of 37 different classes. 
The \textit{ADE 20K} dataset (\cite{zhou2017ade20k}) is a large-scale dataset containing $20,\!100$ images from more than $3K$ classes. The dataset is filtered to contain images of outdoor scenes, resulting in a subset of 5000 images. 
\\

\noindent \emph{Strokes Parametrization.} Following the work of \cite{zou2021stylized}, we parametrize the strokes as $s\!=\!(x, \rho, \sigma, \omega)$. The center of the stroke is represented by $x\!=\!(x_x, x_y)$. 
The height and width of the strokes are represented by $\sigma\!=\!(\sigma_h, \sigma_w)$, while $\omega$ represents the orientation, which is the counter-clock wise angle in the range $[0, \pi]$. 
Lastly, the color of the stroke is represented by $\rho\!=\!(\rho_r, \rho_g, \rho_b)$.
All the parameters are normalized to lie in the interval $[0, 1]$. \\

\noindent \emph{Decomposition.}~We make use of Stylized Neural Painting (SNP) \cite{zou2021stylized}, to extract a sequence of brushstrokes from a given image. We notice that SNP tends to produce very large strokes in the first iterations of the method, which cover a wide area of the canvas. This practice is unrealistic since the size limitations of physical brushstrokes would prevent a human painter from doing this. To circumvent such behavior, we clamp the parameter $\sigma$ to a maximum value of $0.4$.
As described in \cite{zou2021stylized}, we employ a progressive rendering pipeline with a total of 4 iterations, dividing the image in a grid with 4, 9, 16, and 25 regions. We allocate a different number of strokes during the progressive rendering process, respectively 30, 20, 15, and 10 to each region, which results in a total of 790 strokes per image. 
Lastly, SNP represents the color of each stroke using two triples of $(\mathrm{r}, \mathrm{g}, \mathrm{b})$ values that are interpolated to obtain a smooth color. For simplicity, we use the average of the two, and represent the stroke with a uniform color $\rho$. \\

\noindent \emph{Reordering.} We  perform a reordering of the sequences, by minimizing the cost function described in  Sec.\ref{sec:dataset} of the main paper. \\

\noindent \emph{Rendering.} To render the strokes on the canvas, we follow \cite{liu2021painttransformer} and use a \emph{parameter free} renderer. Starting from a primitive brushstroke, affine transformations are applied to obtain the foreground $I_{s^t}$ and the alpha matte $\alpha_{s^t}$ associated to $s^t\!=\!(x, \rho, \sigma, \omega)$. The canvas can be updated computing $I_c^{t}\!=\alpha_{s^t} \cdot I_{s^t} + (1 - \alpha_{s^t}) \cdot \!I_c^{t-1}$. We refer the reader to \cite{liu2021painttransformer} for additional details.

\subsection*{Method}
\label{sec:supp_method}
\noindent \emph{Architecture.} We report more details of the architecture, depicted in Fig. 2 of the main paper. 
Our model relies on the Transformer architecture of \cite{vaswani2017attention}, where we set the embedding dimensionality $\mathrm{d_{emb}}\!=\!256$, the number of heads in multi-head attention to 4, the dimension of the intermediate linear layer to 1024, and the dropout rate to 0.
The CNN encoder $F$ is composed of 4 convolutional blocks with residual connection and receives as input an image of size $256\times256$. The spatial resolution of the features is reduced by a factor of 2 in each block, resulting in a $16\times16$ output feature map. Following \cite{liu2021painttransformer}, we use two distinct image encoders for $I_\mathrm{ref}$ and $I_c$. The features obtained by the two input images are concatenated and projected to the embedding dimensionality $\mathrm{d_{emb}}$. 
Similarly, the context strokes $s_c$ and the target strokes $s_t$ are projected to $\mathrm{d_{emb}}$ using a linear layer. 
The remaining components are implemented as standard transformers blocks. In particular, $C_e$ is transformer encoder with number of layers equal to 8, while $E$, $D_1$, $D_2$ are transformer decoders with number of layers equal to 6. \\

\noindent \emph{Losses.}
We provide additional details about the computation of the losses.
The reconstruction loss component of $\mathcal{L}_{\beta\mhyphen\mathrm{VAE}}$ is computed by weighting the reconstruction error differently for each component of the stroke parameter:

\begin{equation}
\begin{split}
    \left\|s_t - \hat{s}_t\right\|_2^2 & = \lambda_x \left\|x_t - \hat{x}_t\right\|_2^2 + \lambda_\rho \left\|\rho_t - \hat{\rho}_t\right\|_2^2 \\
     & + \lambda_\sigma \left\|\sigma_t - \hat{\sigma}_t\right\|_2^2 + \lambda_\omega \left\|\omega_t - \hat{\omega}_t\right\|_2^2
\end{split}
\end{equation}
with $\lambda_{x}\!=\!1$, $\lambda_{\rho}\!=\!2.5\mathrm{e}{-1}$, $\lambda_{\sigma}\!=\!1$ and $\lambda_{\omega} \!=\!1$. In early experiments, we noticed that the component corresponding to the predicted color $\rho$, \emph{i.e.} $\left\|\rho_t - \hat{\rho}_t\right\|_2^2$, was difficult to jointly optimize with $\mathcal{L}_{\mathrm{col}}$, hence we reduced its weights until convergences of the two.
The last component of our objective is the distribution matching loss $\mathcal{L}^{reg}_{\mathrm{dist}}$. We noticed that, when this loss is used, the predicted strokes may present a distorted height/width ratio. To avoid this issue, when computing this loss we exclude the size $\sigma$ and the orientation $\omega$ from the computation of features $\psi$. The same modification is applied to the SNP+ baseline for fairness of comparison.

\subsection*{Experiments}
\label{sec:supp_experiments}
\noindent \emph{User study.}  We now provide details on the user study presented in the main paper. Each task of the user study consists of an HTML page divided into two sections. In the first, called the demonstration section, we show a collection of stroke sequences taken from the training set. In the second section, called the evaluation section,  we show the users two stroke sequences produced from the same reference images and stroke context, one produced with \methodname~and the other with one of the baselines. We asked the participants to select which of the two sequences of strokes presents characteristics (in terms of stoke positions, colors, and subject consistency) that are most similar to the ones of the strokes in the demonstration section.
To ease the evaluation, we produce sequences with a length of 24 strokes and render the obtained strokes in a short video.
The user study was conducted on 40 images taken from the test set of \textit{Oxford-IIIT Pet INP} dataset, from which a total of 120 tasks was generated. We collected a total of 960 votes from 8 unique users. Examples of the user study are provided on the supplementary website.